\documentclass[letterpaper]{article} 
\pdfoutput=1
\usepackage{aaai23}  
\usepackage{times}  
\usepackage{helvet}  
\usepackage{courier}  
\usepackage[hyphens]{url}  
\usepackage{graphicx} 
\usepackage{natbib}  
\usepackage{caption} 
\usepackage{algorithm}
\usepackage{algorithmic}
\usepackage{amsmath, amssymb, mathtools}
\usepackage{booktabs}
\usepackage{makecell}
\usepackage{multirow}
\usepackage{bbding}
\usepackage{color}
\usepackage{newfloat}
\usepackage{listings}
\DeclareCaptionStyle{ruled}{labelfont=normalfont,labelsep=colon,strut=off} 
\floatstyle{ruled}
\newfloat{listing}{tb}{lst}{}
\floatname{listing}{Listing}
\title{Semantic 3D-aware Portrait Synthesis and Manipulation Based on Compositional Neural Radiance Field}
\author{
    Tianxiang Ma\textsuperscript{\rm 1,2}\equalcontrib,
    Bingchuan Li\textsuperscript{\rm 3}\equalcontrib,
    Qian He\textsuperscript{\rm 3},
    Jing Dong\textsuperscript{\rm 2}\thanks{Corresponding author.},
    Tieniu Tan\textsuperscript{\rm 2,4},
}
\affiliations{
    \textsuperscript{\rm 1}School of Artificial Intelligence, University of Chinese Academy of Sciences\\
    \textsuperscript{\rm 2}CRIPAC \& NLPR, Institute of Automation, Chinese Academy of Sciences\\
    \textsuperscript{\rm 3}ByteDance Ltd, Beijing, China \ \ 
    \textsuperscript{\rm 4}Nanjing University \ \ 
    
    tianxiang.ma@cripac.ia.ac.cn, \{libingchuan, heqian\}@bytedance.com, \{jdong, tnt\}@nlpr.ia.ac.cn,

}

\usepackage{bibentry}

\begin{document}

\maketitle

\begin{figure*}[t]
\centering
\includegraphics[width=1.0\linewidth]{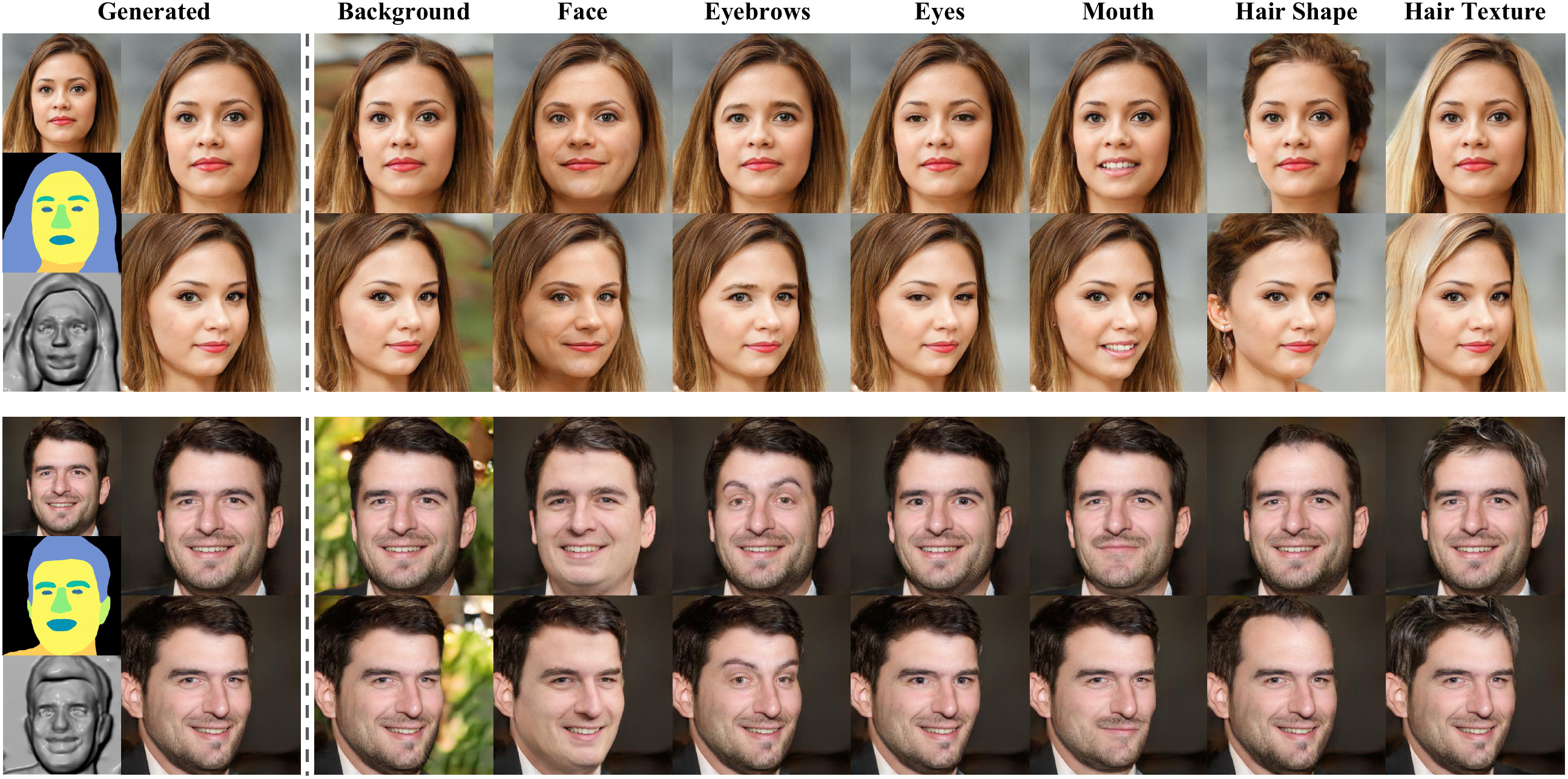}
\caption{The 3D-aware image synthesis and semantic manipulation. The left side shows the randomly generated face images, semantic masks, and 3D shapes. The right side shows the results of manipulating the local semantic regions in different views.}
\label{fig-show}
\end{figure*}

\begin{abstract}
Recently 3D-aware GAN methods with neural radiance field have developed rapidly. However, current methods model the whole image as an overall neural radiance field, which limits the partial semantic editability of synthetic results. Since NeRF renders an image pixel by pixel, it is possible to split NeRF in the spatial dimension. We propose a Compositional Neural Radiance Field (CNeRF) for semantic 3D-aware portrait synthesis and manipulation. CNeRF divides the image by semantic regions and learns an independent neural radiance field for each region, and finally fuses them and renders the complete image. Thus we can manipulate the synthesized semantic regions independently, while fixing the other parts unchanged. Furthermore, CNeRF is also designed to decouple shape and texture within each semantic region. Compared to state-of-the-art 3D-aware GAN methods, our approach enables fine-grained semantic region manipulation, while maintaining high-quality 3D-consistent synthesis. The ablation studies show the effectiveness of the structure and loss function used by our method. In addition real image inversion and cartoon portrait 3D editing experiments demonstrate the application potential of our method. The source code is available: \textcolor{magenta}{https://github.com/TianxiangMa/CNeRF}

\end{abstract}

\section{Introduction}
Photo-realistic image synthesis methods based on 2D Generative Adversarial Networks, such as StyleGAN \cite{karras2019style,karras2020analyzing}, have achieved widespread success. However, they are designed for single-view image generation and do not have the ability to generate 3D-consistent images.

The 3D-aware GAN methods aim to synthesize 3D-consistent images. Many of these methods \cite{sitzmann2019scene,yariv2020multiview,mildenhall2020nerf,schwarz2020graf,niemeyer2021giraffe,chan2021pi} require supervision based on multi-view data or synthesize low-resolution images, which limits the applicability of such methods. Other approaches \cite{gu2021stylenerf,chan2022efficient,or2022stylesdf} explore 3D-consistent image generation under single-view training data, which are inspired by StyleGAN's $w$ latent code modulation network and design similar modulated neural radiance field network. These methods effectively solves the problem that training relies on multi-view data. But they model the whole image as an overall neural radiance field, which limits the semantic part manipulation of synthetic images. \cite{sun2022fenerf} introduces semantic segmentation maps into the rendering process, but it still uses a single rendering network and do not explicitly decouple the local semantic parts. It cannot directly edit the semantic regions of the synthetic image by manipulating the $w$ latent code.

In this paper, we propose the first compositional neural radiance field for semantic 3D-aware portrait synthesis and manipulation, namely CNeRF. It divides the portrait image by semantic regions and learns an independent neural radiance field for each region. Therefore, we can independently control the local semantic regions of the synthetic 3D portrait by manipulating the latent code of the neural radiance field in each semantic region. Specifically, we split the overall generator network into $k$ local semantic 3D generators. Each sub-network consists of modulated MLPs to output feature values, color values, mask values, and residual SDF representation of 3D space for each semantic region. Afterwards, the CNeRF conducts semantic part fusion and volume aggregation to render the complete 2D features, 2D colors and corresponding 2D masks. The 2D features and 2D masks can be further processed by a style-based 2D generator to obtain higher resolution portrait. The whole training process is divided into a low-resolution CNeRF rendering stage and a high-resolution image synthesis stage.

Our method is evaluated mainly on FFHQ dataset and corresponding semantic masks. While generating quality and diversity comparable to state-of-the-arts, our method can also finely manipulate semantic parts of generated portraits, and decouple the shape-texture of the semantic regions. In addition we evaluate our approach on a proprietary cartoon portrait dataset. High-quality generation and semantic part editing of the portraits demonstrate the application potential of our method.

To summarize, our main contributions are as following:

\begin{itemize}

\item {We present the first compositional neural radiance field, which divides the image by semantic regions and learns an independent neural radiance field for each region.}


\item {Based on CNeRF, we propose a semantic 3D-aware portrait synthesis and manipulation method and achieve disentanglement of rendered shapes and textures within each semantic region.}

\item {We propose the global and semantic discriminative loss and shape texture decoupling loss to address semantic entanglement during compositional rendering.}

\item {Experiments reveal that our approach can finely manipulate the semantic parts, while maintaining high-fidelity 3D-consistent portrait synthesis.}

\end{itemize}

\begin{figure*}[t]
\centering
\includegraphics[width=0.96\linewidth]{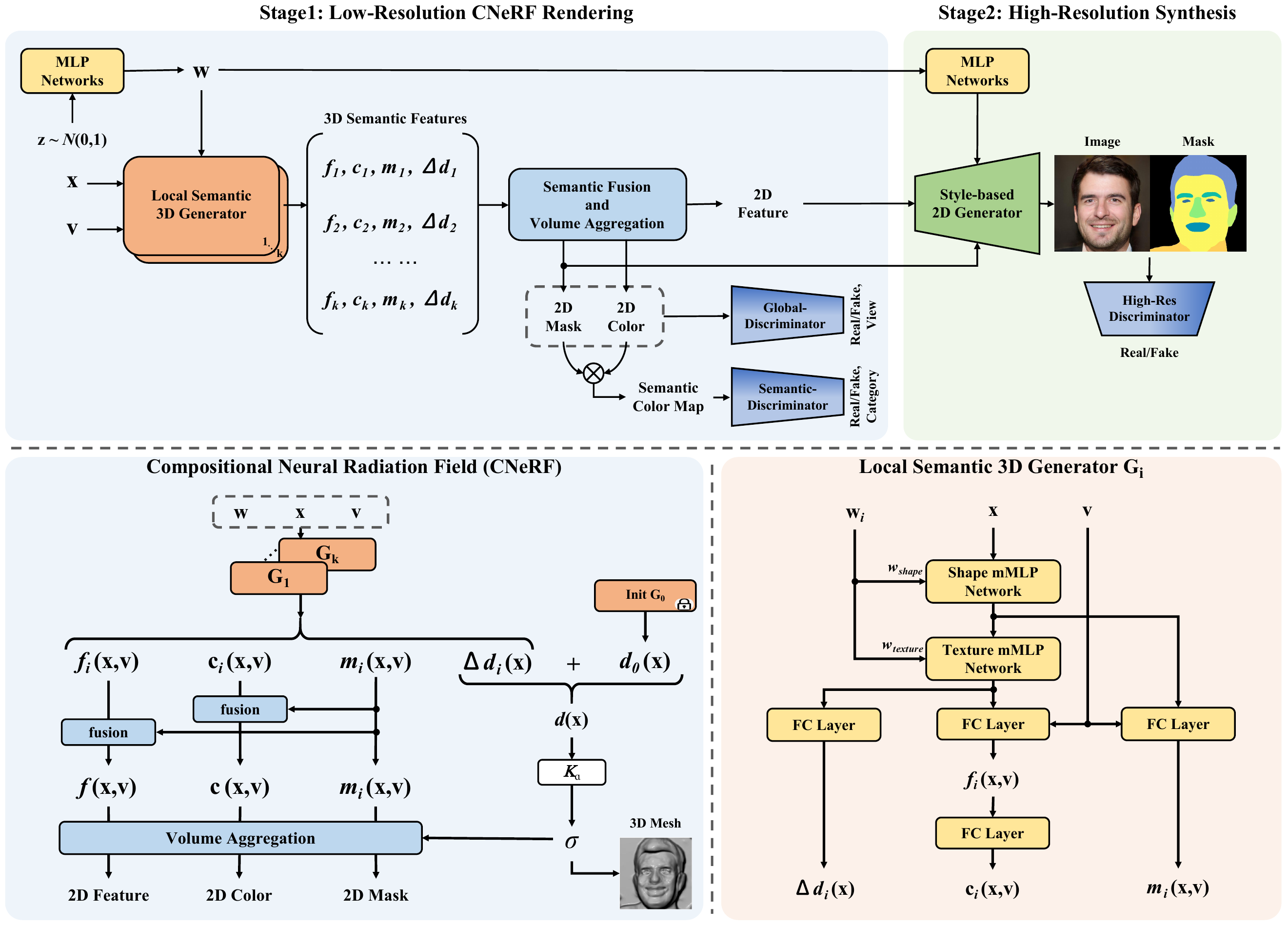}
\caption{The overall framework of our approach. The training process is divided into a low-resolution CNeRF rendering stage and a high-resolution synthesis stage. The specific structure of CNeRF is shown at the bottom left. $k$ local semantic 3D generators learn different semantic part neural radiance field and then sequentially performs semantic-based fusion and volume aggregation. The architectural details of each local semantic 3D generator is shown at the bottom right. The corresponding discriminators and loss functions are used in the two training stages, respectively.}
\label{fig-model}
\end{figure*}

\section{Related Work}
\subsection{2D Generative Adversarial Networks}
2D GANs \cite{goodfellow2014generative} have achieved remarkable success in computer vision. There are many approaches focused on exploring image generation for specific scenarios, including scene generation \cite{park2020swapping,arad2021compositional}, face generation \cite{wiles2018x2face,lee2020maskgan}, human generation \cite{ma2021must,ren2022neural}, etc. One of the most widely used generative frameworks is StyleGAN \cite{karras2019style,karras2020analyzing,karras2021alias}, whose $w$ latent space provides a way to edit different attributes of the generated images. Some approaches explore the editing methods in StyleGAN's latent space, such as InterFaceGAN \cite{shen2020interpreting}, StyleFlow \cite{abdal2021styleflow}, etc. Since StyleGAN is a noise-to-image type of generation method, it cannot input real images for editing. GAN inversion methods are proposed to solve the problem of real image editing using pre-trained StyleGAN, for instance, e4e \cite{tov2021designing}, and PTI \cite{roich2021pivotal}.

\subsection{3D-Aware GANs}
Recently, Neural Implicit Representations (NIR) are widely used for scene 3D modeling \cite{chabra2020deep,chibane2020neural,jiang2020local}, object shape and appearance rendering \cite{michalkiewicz2019implicit,niemeyer2019occupancy,atzmon2020sal,chibane2020implicit,gropp2020implicit}. Some methods \cite{sitzmann2019scene,mildenhall2020nerf,niemeyer2020differentiable} learn neural implicit representation by using multi-view 2D images without 3D data supervision. One of the most important methods is the Neural Radiance Fields (NeRF) \cite{mildenhall2020nerf}, which represents the 3D scene as a series of neural radiance and density fields and uses volume rendering \cite{kajiya1984ray} technique for 3D reconstruction. Many approaches recently focused on 3D-aware GAN under single-view image supervision. Pi-GAN \cite{chan2021pi} proposes a siren-based neural Radiance field and uses global latent code to control the generation of shapes and textures. GIRAFFE \cite{niemeyer2021giraffe} uses a two-stage rendering process, first generating a low-resolution feature map with a volume renderer, and then learning to boost the resolution of the output with a 2D CNN network. StyleNeRF \cite{gu2021stylenerf} integrates NeRF into a style-based generator to improve rendering efficiency and 3D-consistency of high-resolution image generation. EG3D \cite{chan2022efficient} proposes a tri-plane 3D representation method to improve rendering computational efficiency and generation quality. StyleSDF \cite{or2022stylesdf} merges a Signed Distance Fields 3D representation with a style-based 2D generator. Recently, some 3D-aware GAN methods \cite{sun2022fenerf,chen2022sem2nerf} introduce semantic segmentation into the generative network. But they render images and segmentation masks in single network with insufficient semantic part decoupling, which cannot directly edit the semantic regions. In this paper, our method explicitly decomposes the rendering network into multiple semantic region renderers for better semantic decoupling and semantic region editing.

\subsection{Compositional Image Synthesis}
Some methods \cite{arandjelovic2019object,azadi2020compositional,sbai2021surprising} generate complete scenes by combining different images as elements directly at the network input. Other approaches \cite{eslami2016attend,yang2017lr,burgess2019monet,greff2019multi,ehrhardt2020relate,yang2020learning,arad2021compositional,shi2022semanticstylegan} explore the independent implicit representations of different objects in the scene image, which allows to manipulate the objects in the generated scene by controlling the latent code of each object. For 3D-aware GAN methods, it is a novel idea to split the rendered image according to semantic regions and then design a compositional volume rendering method, which allows for individual manipulation of the semantic region in the generated 3D-consistent images.

\section{Methodology}

\subsection{Overview}
The overall model is divided into two stages. Firstly, we train a novel compositional neural radiance field to divide the image by semantic regions and learn an independent neural radiance field for each region, and finally fuses them and renders the complete 2D portrait. Secondly, we utilize a style-based 2D generator to upsample the low-resolution 2D images and obtain high-quality 3D-consistent portraits. In the inference, we can achieve semantic part control of the 3D-consistent portraits by manipulating the latent code of each semantic generator in CNeRF. The overall framework is shown in Figure \ref{fig-model}, and the details are specified below.

\subsection{Compositional Neural Radiance Field}
We propose the Compositional Neural Radiance Field (CNeRF), as shown on the lower left of Figure \ref{fig-model}. We split the overall scene renderer into $k$ local semantic 3D generators, denoted as $G_i$ ($i\in\{1,\dots,k\}$), with $k$ representing the number of semantic categories we set according to the dataset. For the FFHQ dataset we set 9 local semantic categories, i.e., background, face, eyes, eyebrows, mouth, nose, ears, hair and the area below the head. Each local generator consists of modulated Multilayer Perceptron (mMLP) \cite{chan2021pi} and fully connected layers (FC Layers) of the same structure. The generator $G_i$ takes as input the 3D coordinates $\mathbf{x}$, the view direction $\mathbf{v}$, and uses the latent code $w_i$ learned from the noise of the standard Gaussian distribution as the network modulation parameters. $G_i$ outputs two view-dependent color value $c_i(\mathbf{x},\mathbf{v})$ and mask value $m_i(\mathbf{x},\mathbf{v})$, where $c_i(\mathbf{x},\mathbf{v})$ is obtained from the feature output $f_i(\mathbf{x},\mathbf{v})$. In addition $G_i$ outputs a view-independent residual SDF value $\Delta d_i(\mathbf{x})$, where SDF stands for Signed Distance Fields, a proxy of the volume density \cite{yariv2021volume,or2022stylesdf}. The network structure of $G_i$ is shown in the lower right of Figure \ref{fig-model}. 

\noindent \textbf{Local Semantic 3D Generator.}
For each local semantic 3D generator $G_i$, we further decouple the shape and texture properties by splitting the mMLP network into two parts to learn the shape and texture of each semantic part separately. To balance the model performance and the number of parameters, our shape network uses a 3-layer mMLP with SIREN activation \cite{sitzmann2020implicit} and the texture network uses a 2-layer mMLP. The expression for $j$th layer of the mMLP is as follows,
\begin{equation}
\phi_{j}(x)=\sin \left(\gamma_{j}\left(W_{j} \cdot x+b_{j}\right)+\beta_{j}\right),
\end{equation}
where $W_j$ and $b_j$ are network parameters, $\gamma_j$ and $\beta_j$ are modulation coefficients computed from the input $w_i$ ($w_{shape}$ or $w_{texture}$) latent code, and the mapping network for $w_i$ is a 3-layer MLP with LeakyReLU activation. The $w_i$ of each semantic generator may be the same or different at training iteration depending on random settings. Based on the above network, we use the FC layers to output different features at specific layers. Where in addition to the residual SDF value, other outputs are affected by the view direction.

\begin{figure*}[t]
\centering
\includegraphics[width=0.99\linewidth]{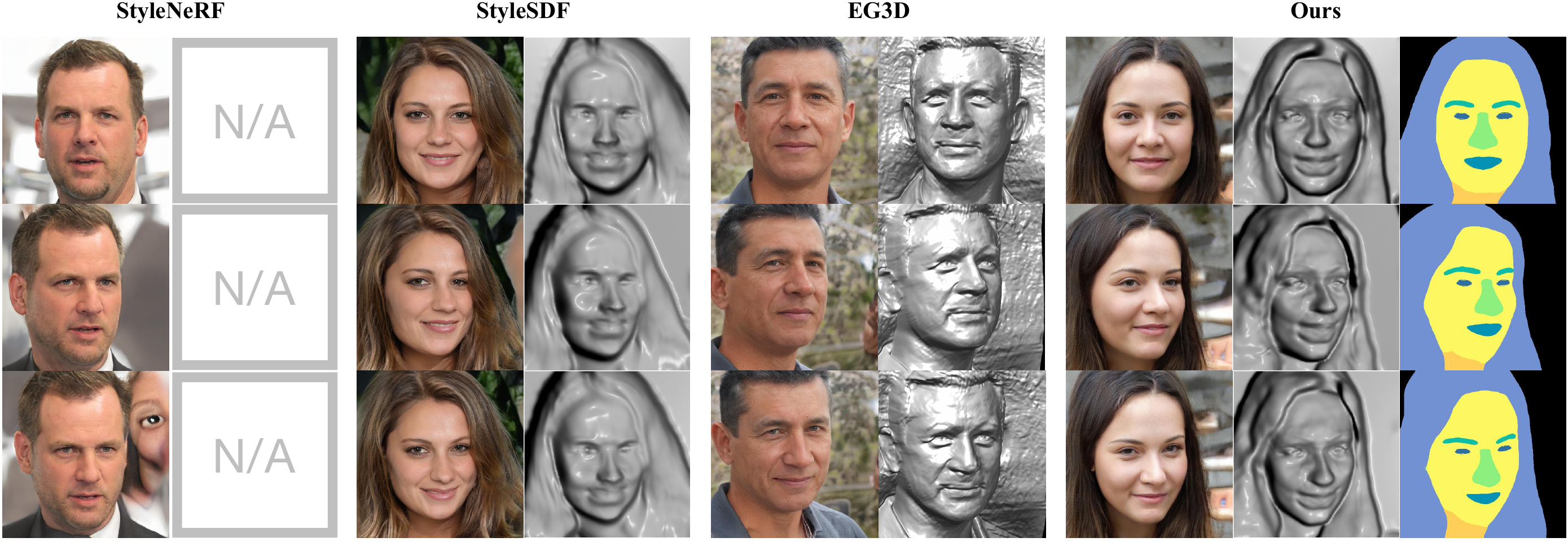}
\caption{Qualitative comparisons of 3D-consistent portrait synthesis between our method and state-of-the-art 3D-aware GAN methods on FFHQ dataset.}
\label{fig-comparison}
\end{figure*}

\noindent \textbf{Residual SDF Representation.}
We tried two types of volume density representation: direct volume density representation and SDF proxy. However, in our experiments we found that splitting the overall generator into local semantic generators destroys the learning process of object 3D shape and leads to incorrect 3D geometric reconstruction. We therefore propose a new volume density proxy, residual SDF representation, for compositional rendering. We add the $\Delta d_i(\mathbf{x})$ output from each semantic generator to a global SDF $d_0(\mathbf{x})$ initialized as a sphere to serve as the true SDF representation for the whole scene. This allows each $G_i$ to learn only the changes of the SDF function value in corresponding region without learning the real physically meaningful SDF value, which makes good 3D geometric reconstruction possible. We transform the residual SDF values of the 3D semantic parts into the density field of the overall scene following the equation below,
\begin{equation}
\begin{aligned}
& \sigma(\mathbf{x})=K_{\alpha}(d(\mathbf{x}))=\frac{1}{\alpha} \cdot \text { Sigmoid }\left(\frac{-d(\mathbf{x})}{\alpha}\right),\\
& d(\mathbf{x})=d_0(\mathbf{x}) + \sum_{i} \Delta d_i(\mathbf{x}), \quad i \in \{1,\ldots,k\},
\end{aligned}
\end{equation}
where $\alpha$ is a learnable parameter to regulate the density tightness of the object surface \cite{or2022stylesdf}. 

\noindent \textbf{Semantic-based Fusion.}
In addition to learning a good global density representation, we need to fuse the color values output by $G_i$ to form a complete color value. We utilize the idea of semantic mask-based fusion. $G_i$ outputs the color $c_i$ and feature $f_i$ along with the mask $m_i$ for the corresponding region, so we use $m_i$ to perform a weighted fusion of $c_i$ and $f_i$,
\begin{equation}
\begin{aligned}
& c(\mathbf{x},\mathbf{v})=\sum_{i} c_i(\mathbf{x},\mathbf{v}) \cdot m_i(\mathbf{x},\mathbf{v}), \quad i \in \{1,\ldots,k\},\\
& f(\mathbf{x},\mathbf{v})=\sum_{i} f_i(\mathbf{x},\mathbf{v}) \cdot m_i(\mathbf{x},\mathbf{v}), \quad i \in \{1,\ldots,k\},
\end{aligned}
\end{equation}
We do not limit the value range of $m_i$ during the semantic fusion, but learn 2D masks end-to-end.

\noindent \textbf{Volume Aggregation.}
For each pixel, we query the samples on a ray that originates at the camera position $\mathbf{o}$, and points at $\mathbf{r}(t) = \mathbf{o} + t\mathbf{v}$. We utilize the classical volume rendering equations to accumulate the samples on the ray,
\begin{equation}
\begin{aligned}
&\mathbf{C}({\mathbf{r}})=\int_{t_{n}}^{t_{f}} T(t) \cdot \sigma(\mathbf{r}(t)) \cdot {c}(\mathbf{r}(t), \mathbf{v}) d t, \\
&\mathbf{F}(\mathbf{r})=\int_{t_{n}}^{t_{f}} T(t) \cdot \sigma(\mathbf{r}(t)) \cdot {f}(\mathbf{r}(t), \mathbf{v}) d t, \\
&\mathbf{M}_i(\mathbf{r})=\int_{t_{n}}^{t_{f}} T(t) \cdot \sigma(\mathbf{r}(t)) \cdot {m}_i(\mathbf{r}(t), \mathbf{v}) d t,
\end{aligned}
\end{equation}
where $T(t)=\exp \left(-\int_{t_{n}}^{t} \sigma(\mathbf{r}(s)) d s\right)$. We use the discrete form of the volume aggregation method following NeRF \cite{mildenhall2020nerf}, and obtain the rendered 2D color, 2D feature and $k$ 2D semantic region masks.

\subsection{Global and Semantic Discriminators}
To help the training of our CNeRF model, we propose two discriminators, the Global-Discriminator (GD), which judges the overall rendering result, and the Semantic-Discriminator (SD), which judges the semantic parts. The GD takes as input the rendered 2D mask and 2D color image and outputs the predicted true/false label as well as the estimated view direction of the input generated image. The estimation of view direction is to ensure a good 3D-consistency of the synthesized image. The SD is designed to enhance semantic region disentanglement. In each iteration of training, the model randomly selects one of the $k$ semantic categories and extracts the corresponding region in the 2D color image based on the generated 2D mask. The SD learns to discriminate the reality of the semantic part input and to predict the semantic category of the input image. Two discriminators  consist of a residual CNN, with the network divided into two output layers at the end. The input of GD has two branches, image CNN and mask CNN. Then the output features of the two branches are added together. Besides, the input of SD is a single branch of semantic region image.

\begin{figure*}[t]
\centering
\includegraphics[width=0.8\linewidth]{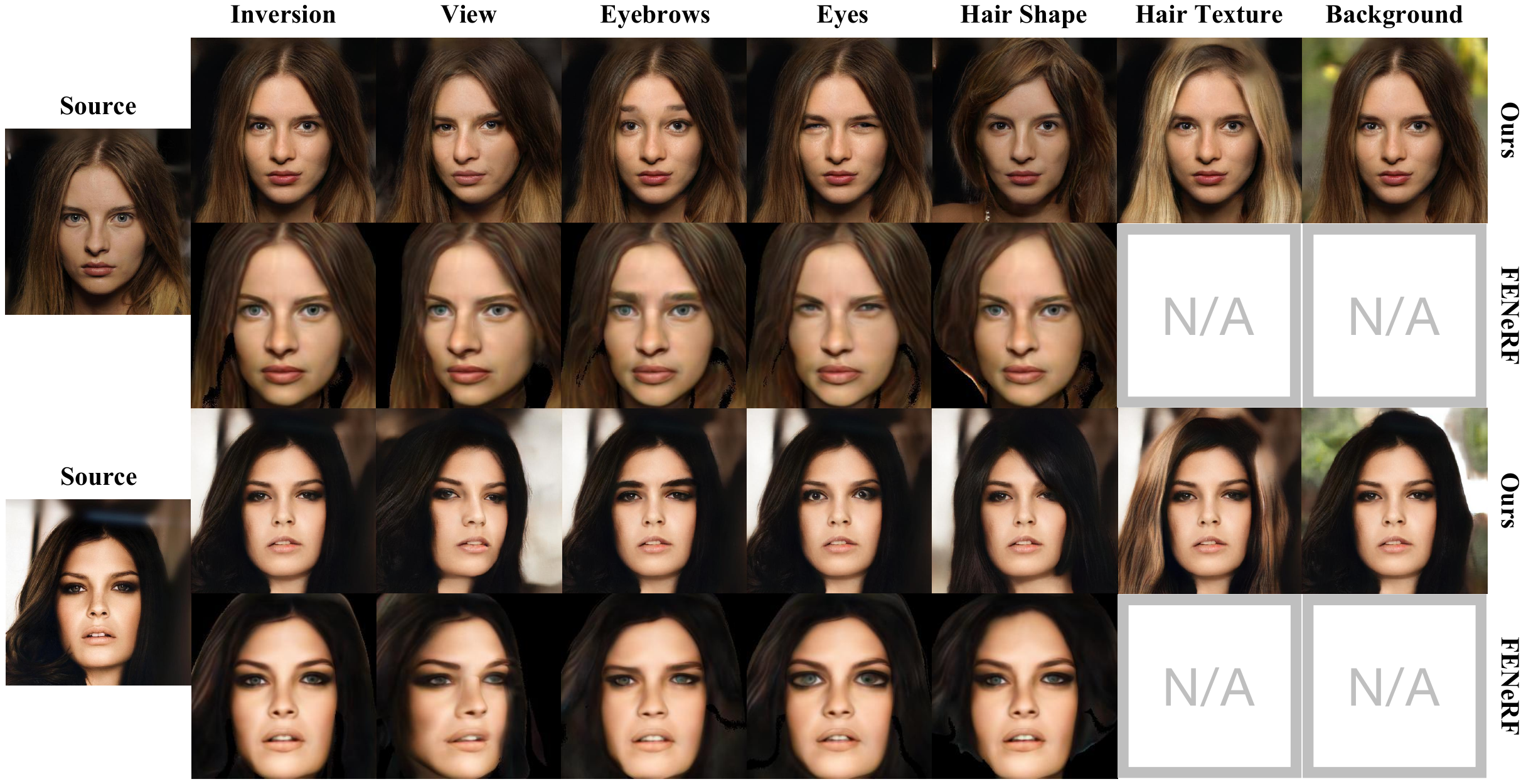}
\caption{Comparisons of image inversion and semantic part manipulation for our method and FENeRF. Training on FFHQ dataset, inversion on CelebA-HQ \cite{liu2015deep} dataset. Perform GAN inversion using PTI \cite{roich2021pivotal}.}
\label{fig-inversion}
\end{figure*}

\subsection{Loss Functions}
\noindent \textbf{Global and Semantic Discriminative Loss.}
For the adversarial losses, we utilize the non-saturating GAN loss with R1 regularization \cite{mescheder2018training} for our two discriminators. denoted as $\mathcal{L}_{adv}^{global}$ and $\mathcal{L}_{adv}^{semantic}$. In addition the two discriminators have additional view direction prediction and semantic category prediction, respectively. We use a smoothed $L1$ loss and a cross-entropy loss to constrain these two learning processes, denoted as $\mathcal{L}_{view}^{global}$ and $\mathcal{L}_{classify}^{semantic}$.

\noindent \textbf{Shape Texture Decoupling Loss.}
Based on our local semantic 3D generator network, we propose a shape texture decoupling loss. It can help better decouple the shape-texture of each semantic region.
\begin{equation}
\begin{aligned}
\mathcal{L}_{STD}&=\left\| \mathcal{M}(w^{1}_{s},w^{1}_{t})-\mathcal{M}(w^{1}_{s},w^{2}_{t}) \right\|_{1} \\ & + max\{ \left\| \phi(I(w^{1}_{s},w^{1}_{t}))-\phi(I(w^{2}_{s},w^{1}_{t})) \right\|_{1} \\ 
& - \left\| \phi(I(w^{1}_{s},w^{1}_{t}))-\phi(I(w^{2}_{s},w^{2}_{t})) \right\|_{1} + m, 0\},
\end{aligned}
\end{equation}
where $w_{s}$ and $w_{t}$ are the abbreviation of $w_{shape}$ and $w_{texture}$.
The first half of the formula indicates the expectation that the generated 2D masks are consistent when we fix the $w_{shape}$ and change the $w_{texture}$. The $\mathcal{M}(*,*)$ represents the rendered 2D mask. The second half represents a contrast loss, where we expect that inputting the same $w_{texture}$ will make the generated 2D color images closer in feature space than inputting two different $w_{texture}$. The $\phi$ is the middle feature layer of the pre-trained VGG network, and $m$ is a fixed margin.

\noindent \textbf{Eikonal Loss and Minimal Surface Loss.} 
Eikonal Loss guarantees that the learned overall SDF of the object is physically valid \cite{gropp2020implicit}.
\begin{equation}
\mathcal{L}_{Eik}=\mathbb{E}_{x}( \left\| \nabla d(x) \right\|_{2} -1)^2.
\end{equation}

Minimal Surface Loss is used to penalize the SDF value to ensure that the surface of the 3D object is smooth and free of false and invisible surfaces \cite{or2022stylesdf}.
\begin{equation}
\mathcal{L}_{MS}=\mathbb{E}_{x}(\exp (- 100 |d(x)|)).
\end{equation}

The overall loss function is,
\begin{equation}
\begin{aligned}
\mathcal{L}_{overall} & =\mathcal{L}_{adv}^{global} + \mathcal{L}_{view}^{global} + \mathcal{L}_{adv}^{semantic} + \mathcal{L}_{classify}^{semantic} \\
& + \mathcal{L}_{STD} + \mathcal{L}_{Eik} + \mathcal{L}_{MS}.
\end{aligned}
\end{equation}
The weights of each loss item are $\lambda_{view}=15$, $\lambda_{Eik}=0.1$, $\lambda_{MS}=0.001$, and the rest are 1.

\subsection{High-Resolution Synthesis}
Direct synthesis of high-resolution 3D-consistent images is difficult, so we use a two-stage training approach \cite{gu2021stylenerf,or2022stylesdf}. First the 64$\times$64 resolution 2D color, 2D feature map, and 2D masks are rendered. Then the CNeRF network weights are fixed and a style-based 2D generator is used to learn higher resolution image synthesis. We utilize the structure of StyleGAN2 \cite{karras2020analyzing} with input 2D feature, and add a mask input branch. We use the CNeRF's latent code $w$ as input to the mapping network of the 2D generator. The 2D generator is shown in the upper right of Figure \ref{fig-model}. We equip the 2D generator with a high-resolution discriminator, which has the same function as the Global-Discriminator of the previous stage.

\begin{table}[t]
\centering
\begin{tabular}{l|cc|ccc}
\bottomrule
 & {FID} & {KID} & {A} & {B} & {C}\\
\hline
{PiGAN (256)}  & {83.0} & {85.8} & {\checkmark} \\
{StyleNeRF (256)}  & {8.0} & {3.7} & {\checkmark} \\
{StyleSDF (256)}   & {11.5} & {2.6} & {\checkmark} \\
{EG3D (512)}       & {4.7} & {0.13} & {\checkmark}  \\
{FENeRF (128)}     & {28.2} & {17.3} & {\checkmark} & {\checkmark}  \\
\hline
{Ours (256)}    & {13.2} & {1.2} & \multirow{2}*{\Checkmark} & \multirow{2}*{\Checkmark} & \multirow{2}*{\Checkmark} \\
{Ours (512)}    & {12.3} & {0.21} \\
\bottomrule
\end{tabular}
\caption{Quantitative evaluation. All methods are evaluated under FFHQ dataset. The resolution of the synthesized image is shown in parentheses. A: 3D-consistent image synthesis, B: Semantic region manipulation, C: Semantic-level shape texture decoupling.}
\label{table:metrics1}
\end{table}

\section{Experiments}

\noindent \textbf{Datasets.}
We train and evaluate our model mainly on the FFHQ \cite{karras2019style} dataset, which contains 70,000 high-quality portrait images with a maximum resolution of 1024$\times$1024. We use the Deeplabv3 \cite{chen2017rethinking} for semantic segmentation of these portrait images. In addition we utilize a 3D rendering pipeline to produce a high-quality cartoon portraits dataset, which contains 20,000 images at 1024 resolution. Similarly we perform facial semantic segmentation on this dataset. Our experimental results are mainly presented in these two datasets at 512 resolution.

\noindent \textbf{Baselines.}
We compare with some of state-of-the-art 3D-aware GAN methods, including PiGAN \cite{chan2021pi}, StyleNeRF \cite{gu2021stylenerf}, StyleSDF \cite{or2022stylesdf}, EG3D \cite{chan2022efficient}, and FENeRF \cite{sun2022fenerf}. Among them, FENeRF is also capable of semantic region editing of 3D-consistent faces, but it renders faces and segmentation masks in single network, which cannot directly edit the semantic regions.

\noindent \textbf{Evaluation Metrics.}
We compute the two most commonly used image quality evaluation metrics, Frechet Inception Distance (FID) \cite{heusel2017gans} and Kernel Inception Distance (KID) \cite{binkowski2018demystifying}, which are used to evaluate the visual quality and diversity of the generated images. There is no unified and accurate evaluation metric for semantic manipulation effect, so we mainly perform qualitative comparison and demonstration.

\noindent \textbf{Implementation Details.}
We train our model in two stages. First, we train CNeRF using 64$\times$64 resolution portrait images and segmentation masks, and render each output at 64$\times$64 resolution. We use ADAM \cite{kingma2014adam} optimizer with learning rates of $2 \times 10^{-5}$ and $2 \times 10^{-4}$ for the generator and discriminator respectively and $\beta_1 = 0, \beta_2 =
0.9$. Second, we freeze the CNeRF weights and train the 2D generator with same setup to StyleGAN2. Our experiments are carried out on 4 32GB Tesla V100 GPUs. Due to limited space more implementation details and detailed architecture of 2D generator and discriminators as well as more experimental results are presented in the supplementary material.

\subsection{Comparisons}
\noindent \textbf{3D-Consistent Portrait Synthesis.}
Figure \ref{fig-comparison} shows the comparative results of 3D-consistent portrait generation. Our method is able to accurately reconstruct the 3D shape of the faces and also generate 2D masks corresponding to the faces. Our face 3D shape reconstruction is comparable to StyleSDF, but our method has semantic part manipulation ability that other comparison methods do not have. Table \ref{table:metrics1} presents the quantitative comparison results of our method with SOTA 3D-aware GAN methods, and the performance of our method is comparable to that of StyleSDF. The main contribution of our approach, however, is to synthesize 3D-consistent images while achieving good semantic-level property manipulation.

\noindent \textbf{Semantic Region Manipulation.}
Figure \ref{fig-show} visually demonstrates the manipulation of the semantic parts of the synthesized faces, while ensuring 3D-consistent image synthesis. Our method can also control the shape and texture of the semantic regions separately, such as hair region in the figure. In addition figure \ref{fig-inversion} shows the comparison results of our method and FENeRF in real image semantic part manipulation. We use PTI \cite{roich2021pivotal} to train GAN inversion. Our method inverts more realistic face images with better semantic part controllability. For example, our method can control the position as well as the shape of the eyebrows, while FENeRF can only edit at the original eyebrow area. Our model is able to decouple semantic region shape and texture properties, such as the displayed hair area, and change the background, which FENeRF cannot do. In addition our method manipulates the semantic parts by directly controlling the $w_i$ latent code of each semantic generator, while FENeRF can only edit the mask manually and perform semantic editing with the help of GAN inversion.

\begin{figure}[t]
\centering
\includegraphics[width=1.0\linewidth]{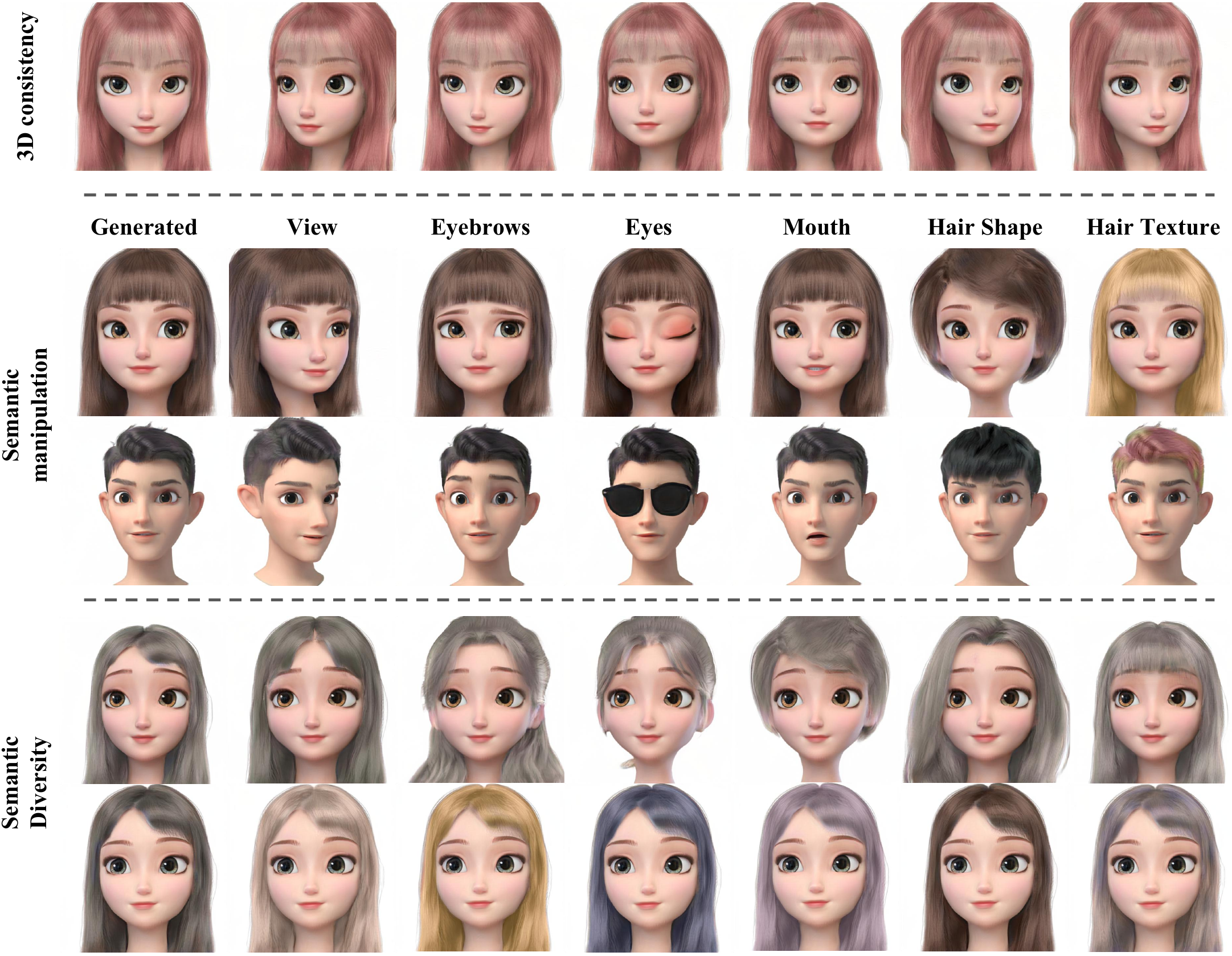}
\caption{The 3D-consistent image synthesis and semantic region editing of cartoon portraits.}
\label{fig-pixar}
\end{figure}

\begin{figure}[t]
\centering
\includegraphics[width=1.0\linewidth]{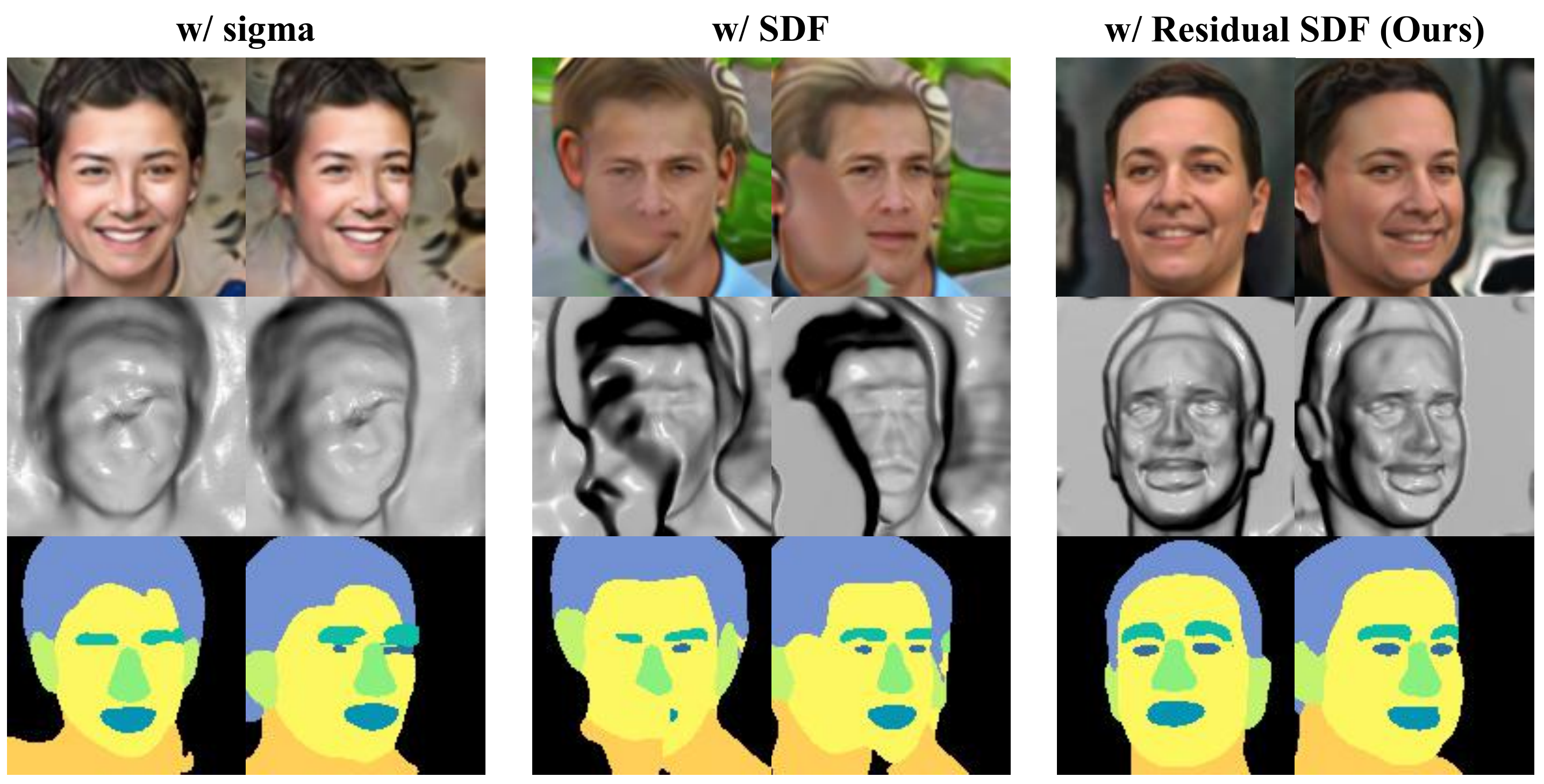}
\caption{Ablation study of density function proxy representation in compositional rendering. Showing 128 resolution rendering results.}
\label{fig-ablation1}
\end{figure}

\begin{figure}[t]
\centering
\includegraphics[width=1.0\linewidth]{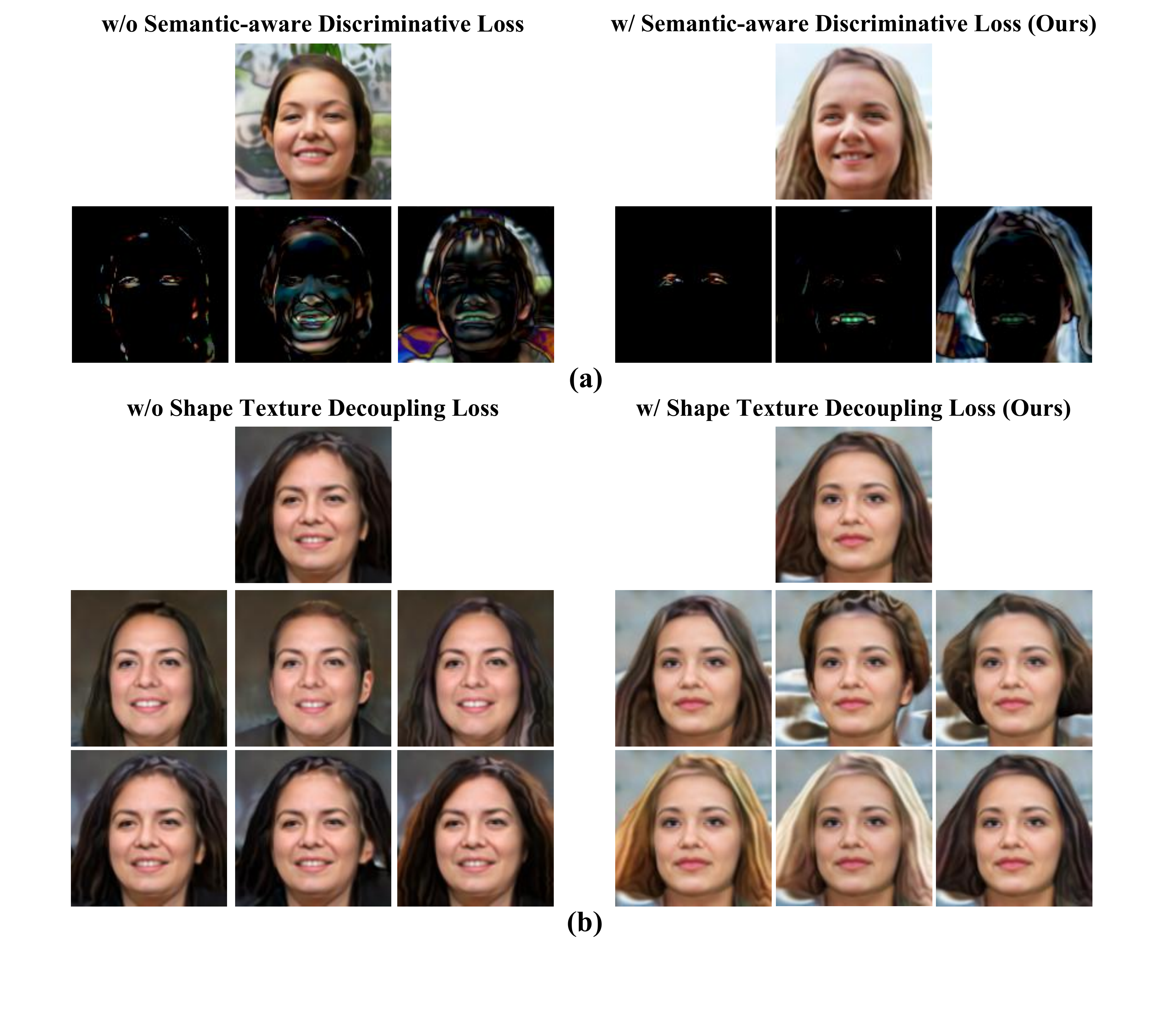}
\caption{(a) Ablation of semantic discriminative loss. The second row represents the residual between the local manipulated image and the original image. (b) Ablation of shape-texture decoupling loss. The last two rows show the manipulation of hair shape and texture.}
\label{fig-ablation2}
\end{figure}

\subsection{Applications}
Our method can be applied to the editing of real images, as shown in Figure \ref{fig-inversion}. The quality of the image inversion and the semantic part manipulation effect illustrate the great potential of our method for application. In addition we test our method on a proprietary cartoon portrait dataset, as shown in Figure \ref{fig-pixar}. High-quality 3D-consistent portrait synthesis as well as accurate and diverse semantic part editing demonstrate the potential of our approach to 3D animation, digital human and metaverse applications.

\subsection{Ablation Studies}
We conduct ablation experiments on our CNeRF model from two aspects. For the overall 3D-consistent image synthesis ability, the residual SDF representation proposed in this paper plays a key role for compositional neural radiance field, as shown in Figure \ref{fig-ablation1}. When using the original volume density or the direct SDF proxy, our CNeRF produces incorrect 3D reconstruction and the synthesized images lose 3D-consistency. For the semantic part decoupling capability, we perform ablation experiments for the two correlated loss functions proposed in this paper, as shown in Figure \ref{fig-ablation2}. When using semantic discriminative loss (and Semantic-Discriminator), there is more accurate control over the semantic parts. When using shape texture decoupling loss, the model's performance is improved on the disentanglement of shape and texture properties and editing diversity.

\section{Conclusion}
In this paper, we propose a novel Compositional
Neural Radiance Field for semantic 3D-aware portrait synthesis and manipulation. It enables high-quality 3D-consistent image synthesis while achieving independent manipulation of the semantic parts and decoupling the shape and texture within each semantic region. Qualitative and quantitative comparison experiments demonstrate the effectiveness and novelty of our method. The application performance of real image inversion and cartoon portrait 3D-aware editing show the potential of the method. There is still room to improve the quality of our approach in 3D reconstruction, and in future work we will explore the combination of our method with the latest 3D-aware GANs, such as EG3D.

\section{Acknowledgments}
This work is supported by the National Key Research and Development Program of China under Grant No. 2021YFC3320103.

\bibliography{aaai23}

\appendix
\newpage

\section{Implementation Details}
\subsection{Datasets Setting}
Since our method requires the use of semantic segmentation maps for training the Compositional Neural Radiance Field, the optimal dataset is the human face dataset.
We use the widely adopted and largest face dataset FFHQ, which has rich head views and facial styles, and with the help of mature face segmentation algorithms, we can get accurate facial semantic segmentation masks. On this dataset, we train our model with a fixed camera field of view at $12^\circ$, and the azimuth
and elevation angles are sampled in a Gaussian distribution with mean 0 and standard deviations of 0.3 and 0.15. In addition we set the near and far fields to [0.88, 1.12] and sample 24 points per ray during training. To further validate the application value of our method, we also conduct experiments on a cartoon-style portrait dataset. The camera parameters for the experiments are set the same as FFHQ, except that the standard deviations of azimuth and elevation angles are set to 0.4 and 0.2. The R1 regularization weight of the image branch in the GAN training loss is 10, and the weight of the mask branch is 1000.

\subsection{Network Details}

\noindent \textbf{Local Semantic 3D Generator.}
The network parameters of each local semantic 3D generator in our Compositional Neural Radiance Field: the number of channels per network layer is 128, the length of the modulation code $w_i$ is 256, and the depths of the shape network and texture network are 3 and 2, respectively. Each FC layer has a layer number of 1. The output $f_i$ has 128 channels, $c_i$ has 3 channels, and $m_i$ and $\Delta d_i$ have 1 channel. 

The latent code $w_i$ of each semantic generator at training time is from the same mapping network, which has a depth of 3 and 256 channels. In order to make the model better distinguish the latent code $w_i$ of different semantics, we adopt a random sampling mechanism: Two $w$ are sampled from the mapping network at each iteration, and each local semantic 3D generator randomly selects one of these $w$ as its own latent code $w_i$. This approach makes the $w_i$ is different for semantic generators at each iteration, but the difference is not so great to make the model difficult to learn.

\begin{figure*}[ht]
\centering
\includegraphics[width=0.6\linewidth]{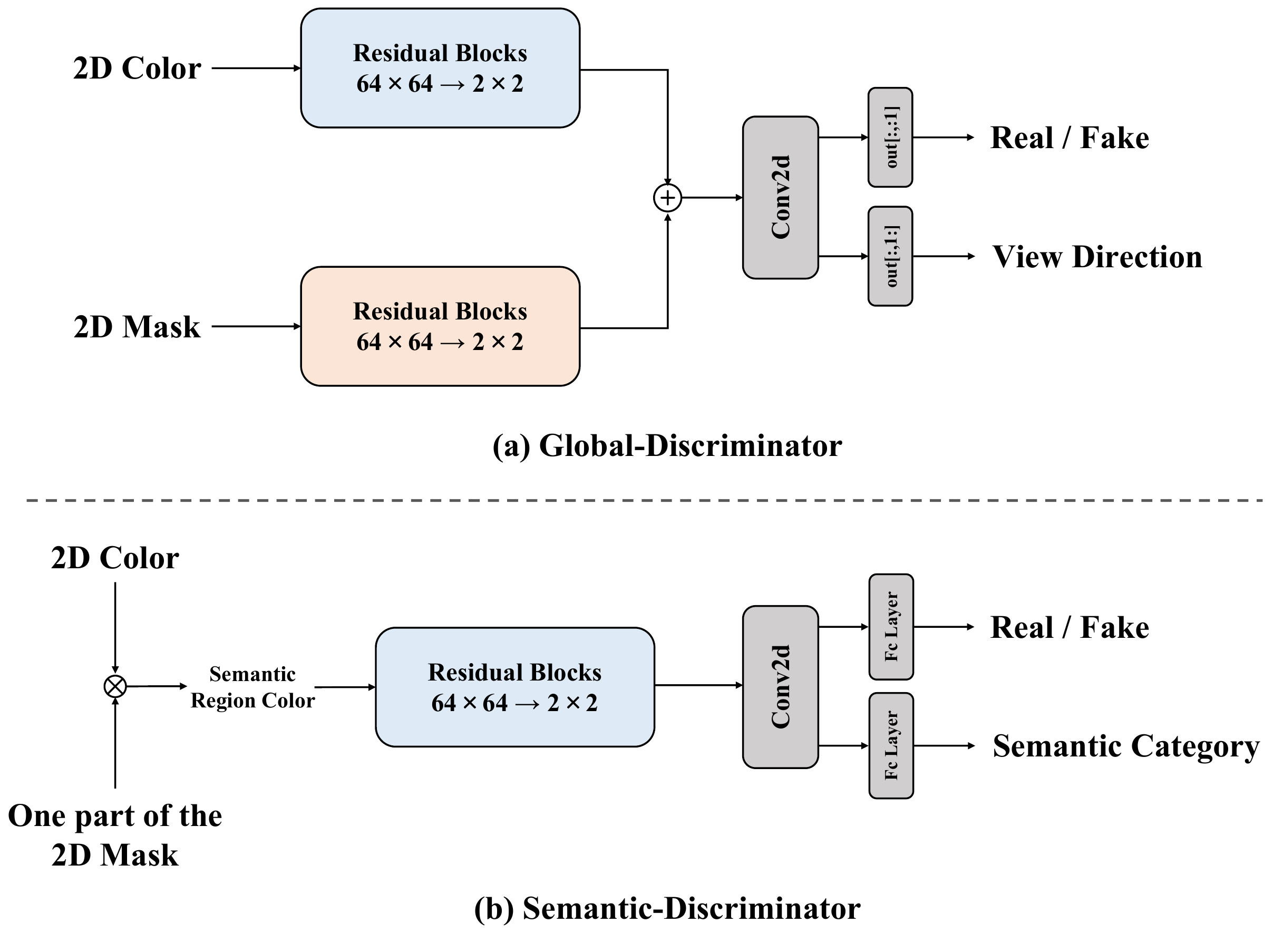}
\caption{The detailed architecture of Global-Discriminator and Semantic-Discriminator.}
\label{fig:discriminator}
\end{figure*}

\noindent \textbf{Global and Semantic Discriminators.}
The detailed architecture of our Global and Semantic Discriminators is shown in Figure \ref{fig:discriminator}. The backbones of the two networks consist of convolutional residual blocks. The Global Discriminator has two branches, while the Semantic Discriminator has one branch. The output value of the Global Discriminator is directly split into two parts to get the True/False label and the estimated view direction, respectively. Note that the network outputs the predicted view direction only when the input image is a generated image, because there are camera parameters that were randomly sampled during training, while the real data do not have corresponding true values of the parameters. The Semantic Discriminator's output layer uses two Fully Connected layers to obtain the True/False label and the predicted input semantic region category, respectively.

\begin{figure*}[t]
\centering
\includegraphics[width=0.6\linewidth]{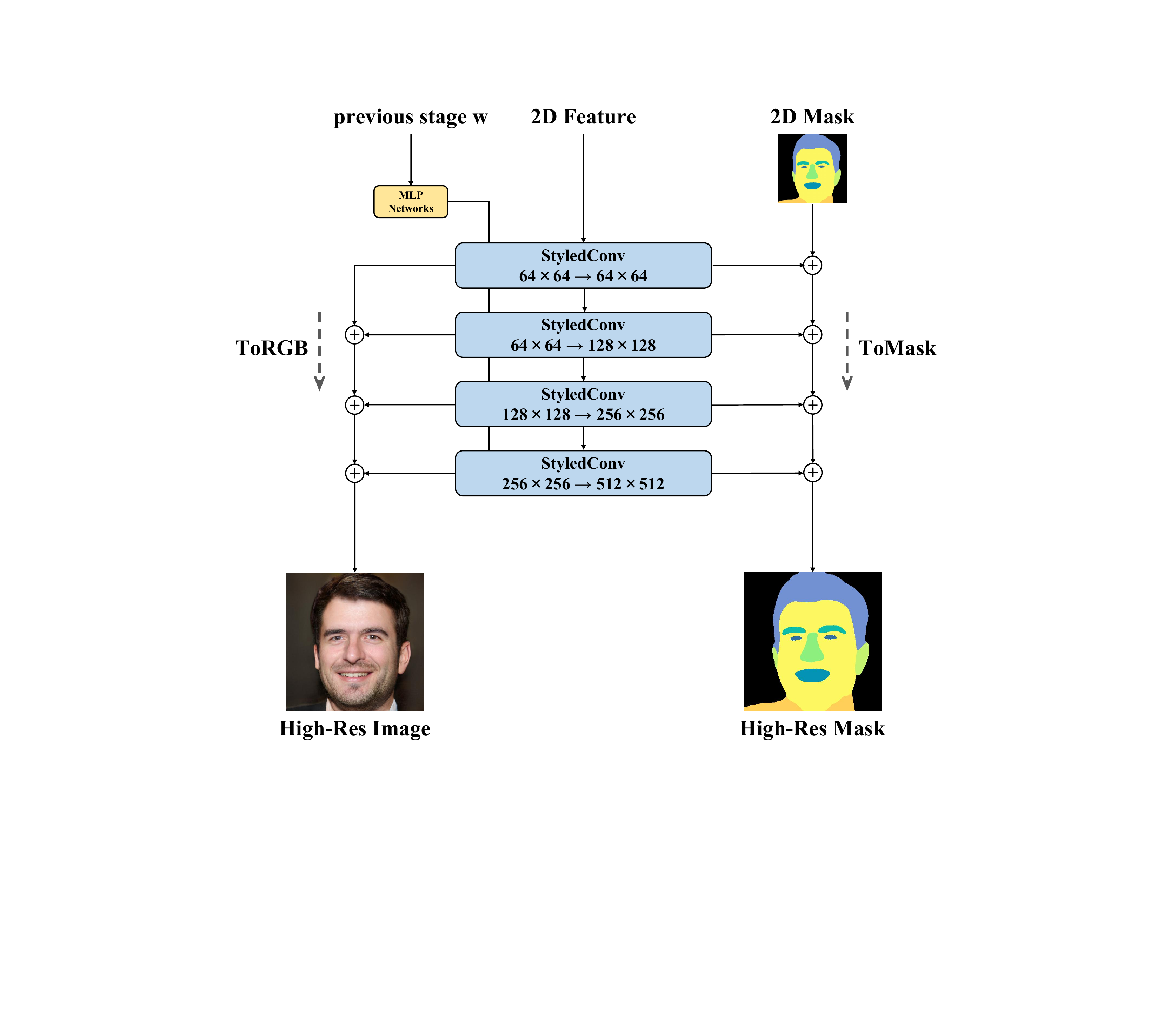}
\caption{The detailed architecture of Style-based 2D Generator.}
\label{fig:generator2D}
\end{figure*}

\noindent \textbf{Style-based 2D Generator.}
The detailed architecture of our style-based 2D Generator is shown in Figure \ref{fig:generator2D}. The input of the network is 2D Feature from CNeRF, and the structure of each StyledConv layer is the same as StyleGAN2, but with two output branches, the image branch and the mask branch. The input of mapping network for this network is from previous stage.

\section{Additional Experimental Results}

\noindent \textbf{Semantic Manipulation Method.}
The direct effect of our approach is that we can individually manipulate the semantic regions of the generated images while ensuring the synthesis of 3D-consistent images. The simplest method is to directly control the latent code $w_i$ of the local semantic 3D generator of the trained model. Thanks to our Compositional Neural Radiance Field, when we manipulate the $w_i$ in one of the semantic regions while fixing the $w_i$ in other parts, only that semantic region changes in the synthetic image.

\begin{figure*}[t]
\centering
\includegraphics[width=0.8\linewidth]{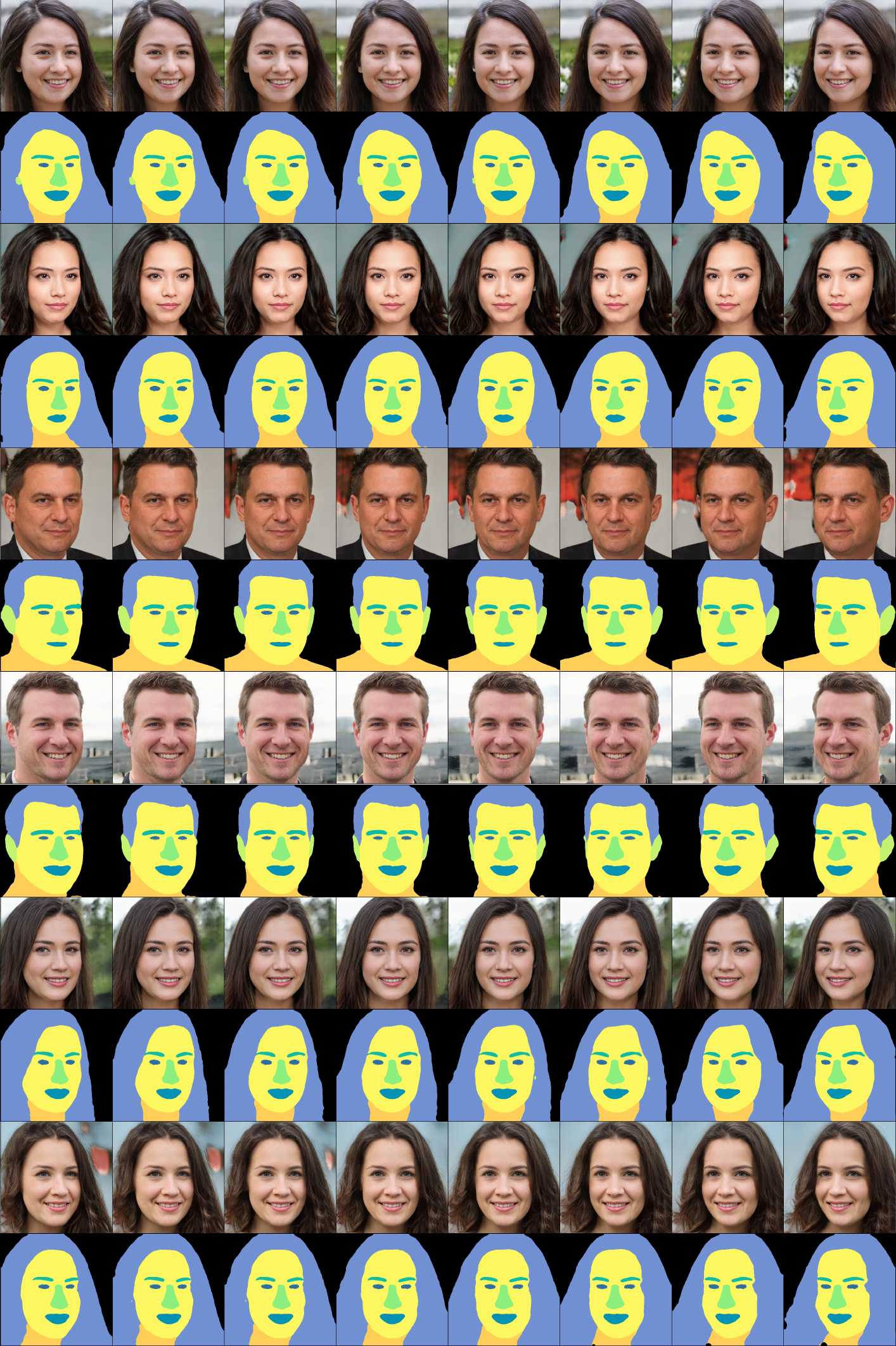}
\caption{The 3D-consistent image and mask synthesis of our method at 512 resolution on the FFHQ dataset.}
\label{fig:supp_continue_views}
\end{figure*}

\noindent \textbf{3D-consistent Image Synthesis Results.}
Figure \ref{fig:supp_continue_views} shows the effect of our 3D-consistent image synthesis at 512 resolution on the FFHQ dataset. Our method is capable of synthesizing high-resolution 3D view-consistent face images and semantic masks.

\begin{figure*}[t]
\centering
\includegraphics[width=0.9\linewidth]{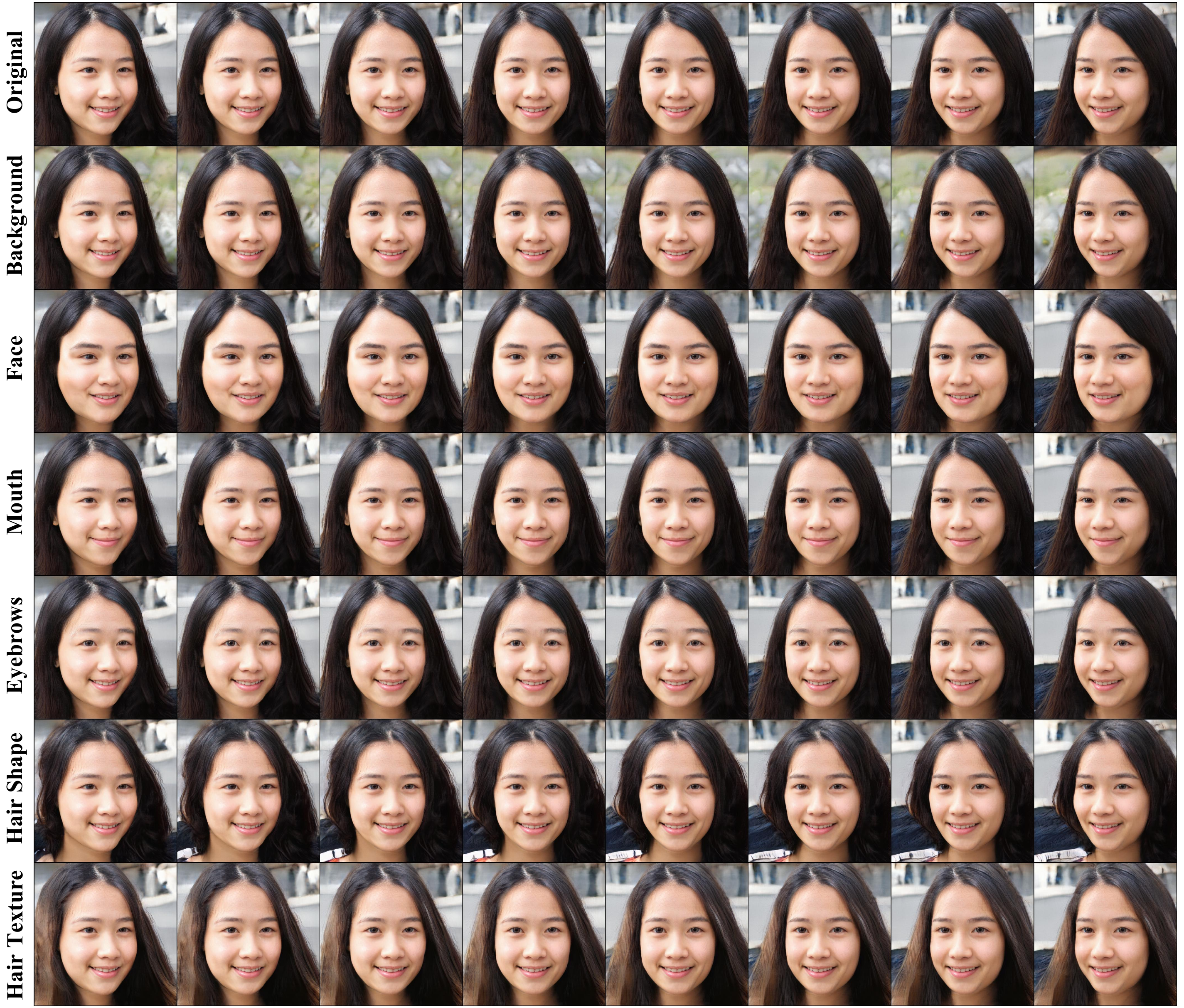}
\caption{The 3D-consistent image semantic region manipulation of our method on FFHQ dataset.}
\label{fig:supp_ffhq_edit1}
\end{figure*}

\begin{figure*}[t]
\centering
\includegraphics[width=0.9\linewidth]{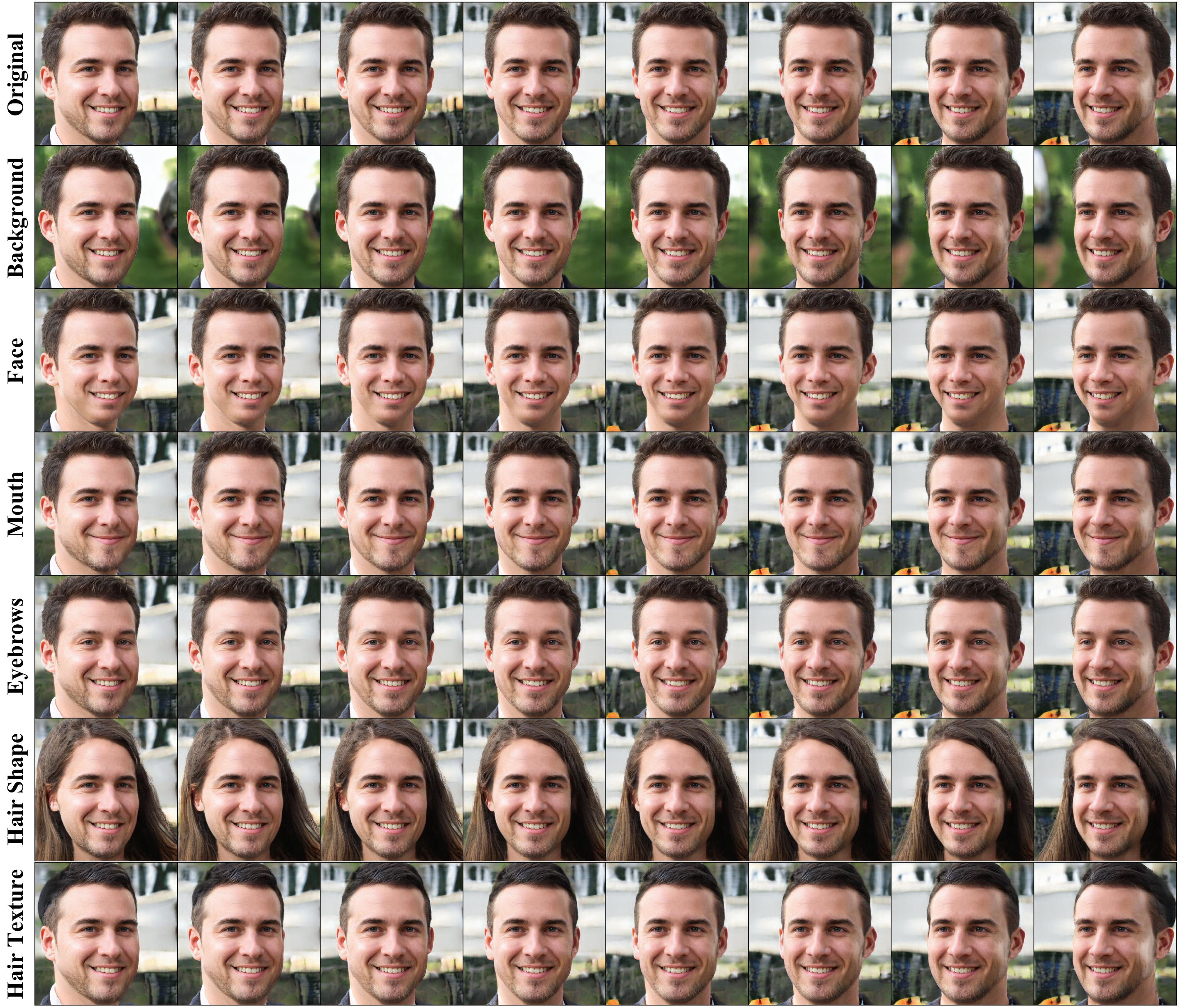}
\caption{The 3D-consistent image semantic region manipulation of our method on FFHQ dataset.}
\label{fig:supp_ffhq_edit2}
\end{figure*}

\begin{figure*}[t]
\centering
\includegraphics[width=0.9\linewidth]{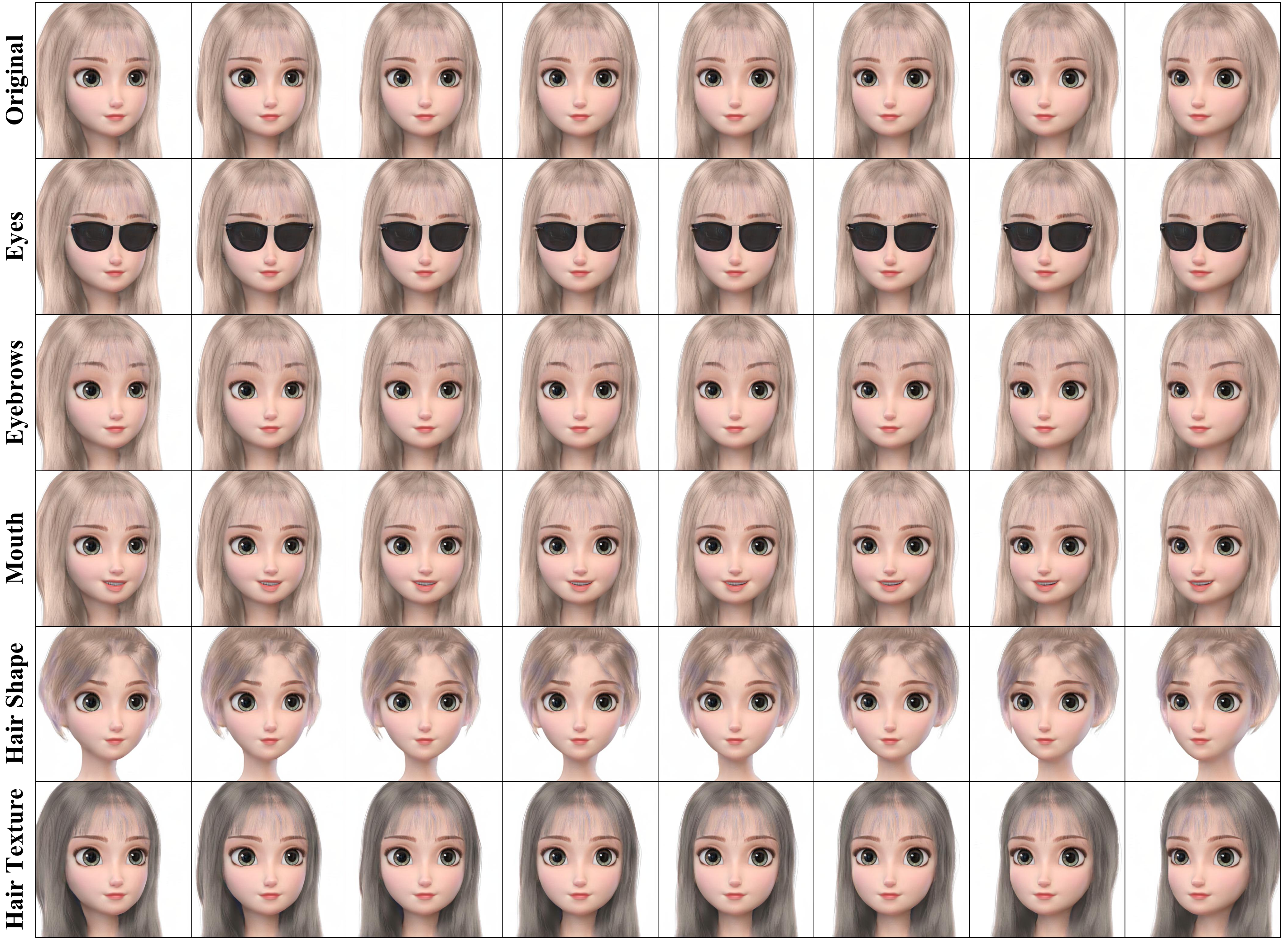}
\caption{The 3D-consistent cartoon portrait synthesis and semantic region manipulation of our method.}
\label{fig:supp_pixar_edit1}
\end{figure*}

\begin{figure*}[t]
\centering
\includegraphics[width=0.9\linewidth]{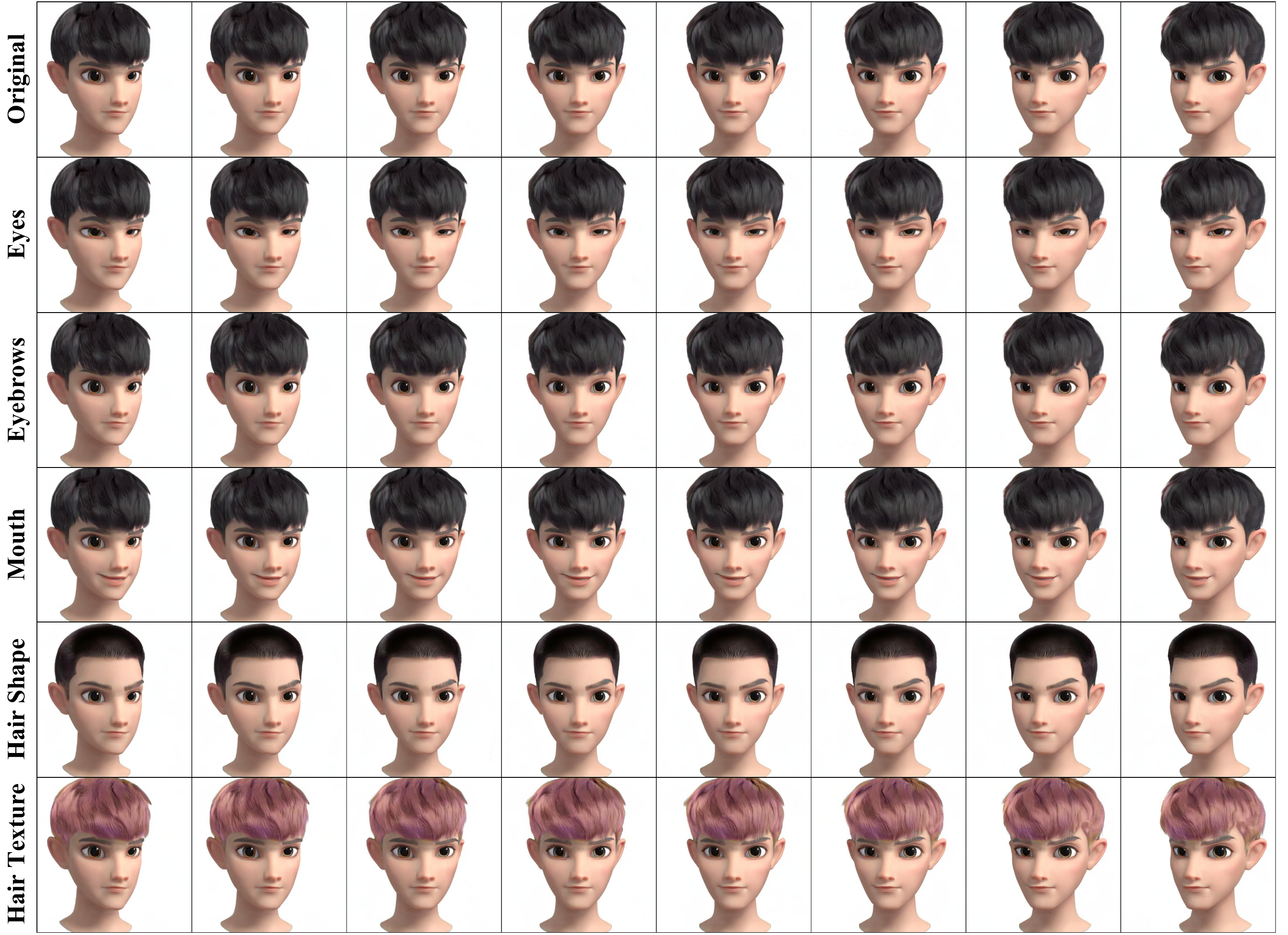}
\caption{The 3D-consistent cartoon portrait synthesis and semantic region manipulation of our method.}
\label{fig:supp_pixar_edit2}
\end{figure*}

\noindent \textbf{Semantic Region Manipulation of Synthetic Images.}
Figure \ref{fig:supp_ffhq_edit1} and \ref{fig:supp_ffhq_edit2} show the effect of semantic region editing for the synthetic images on the FFHQ dataset. The first row is the original synthetic image, the remaining rows are the results of editing the corresponding semantic regions. The last two rows are the results of editing the shape and texture of the hair region, since the most obvious way to show the effect of shape texture decoupling is to use the hair area. In addition Figure \ref{fig:supp_pixar_edit1} and \ref{fig:supp_pixar_edit2} show the effect of our method on cartoon-style portrait 3D-consistent synthesis and semantic region editing.

\begin{figure*}[t]
\centering
\includegraphics[width=0.9\linewidth]{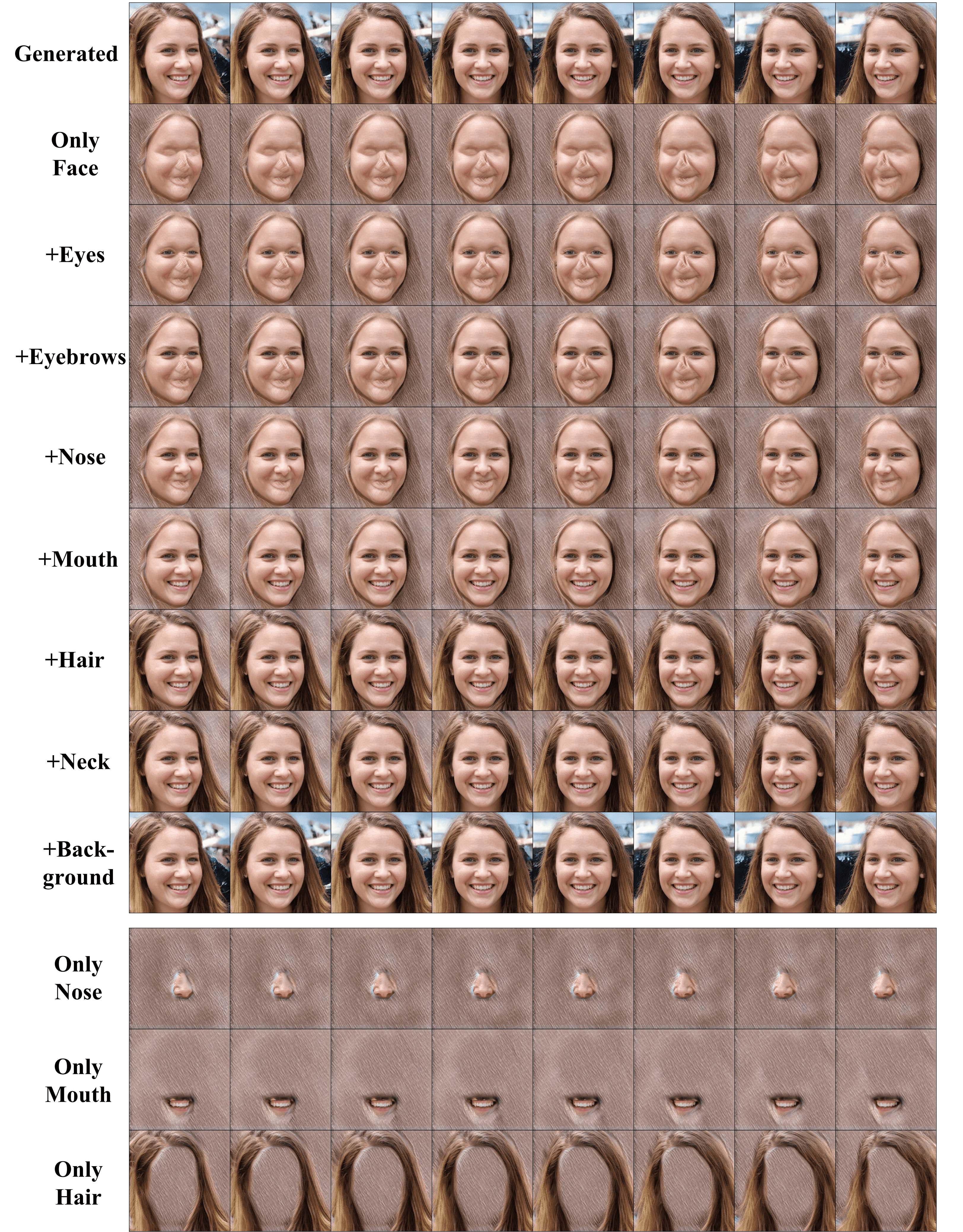}
\caption{Independent semantic region synthesis results of our method when individual or partially fused semantic region features are output in CNeRF.}
\label{fig:independent}
\end{figure*}

\noindent \textbf{Independent Semantic Region Synthesis}
Our approach is able to generate local semantic parts independently when no fusion is performed. As shown in Figure \ref{fig:independent}, when our method does not use semantic fusion but outputs a certain semantic region feature alone in CNeRF, the final synthesis effect only outputs the corresponding semantic region. At the same time, synthetic faces can still ensure 3D consistency. Gradually increase the semantic region features, and the synthetic face will also gradually increase the corresponding region attributes. This model feature will provide more application feasibility.

\end{document}